\documentclass{article}
\usepackage{microtype}
\usepackage{graphicx}
\usepackage{subfigure}
\usepackage{booktabs}
\usepackage{enumitem}
\usepackage{wrapfig}
\usepackage{caption}
\usepackage{subcaption}
\usepackage[authoryear,round]{natbib}  
\usepackage{hyperref}
\usepackage{amsmath}
\usepackage{amssymb}
\usepackage{mathtools}
\usepackage{amsthm}
\usepackage[capitalize,noabbrev]{cleveref}
\usepackage[english]{babel}
\usepackage[margin=1in]{geometry}
\usepackage[utf8]{inputenc}
\usepackage[T1]{fontenc}
\usepackage{microtype}
\usepackage{xcolor}

\title{Transformers Don't In-Context Learn Least Squares Regression}
\author{
  Joshua Hill$^{1}$\thanks{Corresponding author: \texttt{jhdhill@uwaterloo.ca}}
  \and
  Benjamin Eyre$^{2}$
  \and
  Elliot Creager$^{1,3}$
}
\date{}

\newcommand\extrafootertext[1]{%
    \bgroup
    \renewcommand\thefootnote{\fnsymbol{footnote}}%
    \renewcommand\thempfootnote{\fnsymbol{mpfootnote}}%
    \footnotetext[0]{#1}%
    \egroup
}

\begin{document}
\maketitle

\extrafootertext{
This paper was accepted (Oral Presentation) at the ICML 2025 Workshop on Reliable and Responsible Foundation Models}

\vspace{-2em} 
\begin{center}
  $^{1}$University of Waterloo \ 
  $^{2}$Columbia University \ 
  $^{3}$Vector Institute
\end{center}

\begin{abstract}

In-context learning (ICL) has emerged as a powerful capability of large pretrained transformers, enabling them to solve new tasks implicit in example input–output pairs without any gradient updates. Despite its practical success, the mechanisms underlying ICL remain largely mysterious. In this work we study synthetic linear regression to probe how transformers implement learning at inference time.
Previous works have demonstrated that transformers match the performance of learning rules such as Ordinary Least Squares (OLS) regression or gradient descent and have suggested ICL is facilitated in transformers through the learned implementation of one of these techniques.
In this work, we demonstrate through a suite of out‑of‑distribution generalization experiments that transformers trained for ICL fail to generalize after shifts in the prompt distribution, a behaviour that is inconsistent with the notion of transformers implementing algorithms such as OLS.
Finally, we highlight the role of the pretraining corpus in shaping ICL behaviour through a spectral analysis of the learned representations in the residual stream. Inputs from the same distribution as the training data produce representations with a unique \textit{spectral signature}: inputs from this distribution tend to have the same top two singular vectors. This spectral signature is not shared by out-of-distribution inputs, and a metric characterizing the presence of this signature is highly correlated with low loss.

\end{abstract}

\section{Introduction}
\label{sec:intro}

Deep learning models, particularly transformers \citep{trans}, have shown impressive performance across various tasks. In addition to state-of-the-art performance on tasks on which these models are explicitly trained on, large transformers trained on variations of the language modelling objective (LLMs) demonstrate the perplexing ability to perform previously unseen tasks using a small sequence of input-output pairs as context. This ability, known as In-Context Learning (ICL), has helped propel LLMs into popular use by supplying an easy to use interface for a machine learning system capable of several tasks with no additional training~\citep{brown2020,dong2024survey}.

Despite its importance to modern machine learning, relatively little is understood about how ICL works and which aspects of transformer training enable this behavior. 
This has prompted recent investigations into the limitations of this ability, the circumstances in which it can emerge, and the mechanisms that can facilitate learning at inference time. 
Notably, several works have used the problem of in-context regression using synthetic data as a controlled test bed for developing and evaluating ICL hypotheses \citep{garg2023, goddard2024specializationgeneralization, akyurek2023}. Within this setting, several theories have been posited suggesting that training a transformer on this task produces a model which naturally implements known regression learning rules like Ordinary Least Squares (OLS) or gradient descent, and that this is the mechanism enabling ICL \citep{akyurek2023, von2023transformers, garg2023}.

Following recent re-evaluations of these hypotheses \citep{arora2024bayesian, shen2023pretrained}, we perform experiments to critique existing theories and motivate new ones. In particular, we provide evidence for the following claims:

\begin{enumerate}
    \item OLS regression, gradient descent, and other similar regression rules are incapable of explaining the generalization performance of transformers performing ICL. 
    \item The ability for ICL to be performed effectively relies heavily on the \emph{pre-training data} and its relation to the in-context examples, and performance can be predicted by spectral signatures in the geometry of the residual stream representation of the prompt.
\end{enumerate}

We support our first claim through out-of-distribution generalization experiments where the transformer is trained on a series of ICL regression tasks, where the training set bears a systematic difference from the test set. While traditional regression rules guarantee a reasonable level of generalization regardless of the subspace the inputs reside in, we find that transformers performing ICL are only able to correctly perform regression on sequences similar to the ones they trained on. Even more surprisingly, we find that even in the in-distribution generalization case, transformers performing ICL achieve an asymptotic error rate orders of magnitude worse than OLS.

Our second claim proposes
that the representation geometry of a prompt is heavily influenced by the prompt's similarity to those seen during pre-training.
By probing residual-stream activations before the final readout head, we uncover distinct patterns in the geometry of the residual stream for in-distribution versus out-of-distribution (OOD) prompts, which we call \textit{spectral signatures}. In-distribution prompts vary within a stable low-dimensional subspace, evidenced by rapidly decaying singular-value spectra and consistent principal directions, while OOD prompts produce flatter, high-variance spectra with unstable principal directions. 

These signatures correlate strongly with performance and the distributional source of the prompt, and are seen across different training distributions.

The widespread adoption of foundation models, and ICL specifically, as a general purpose machine learning pipeline makes characterizing the shortcomings of these models all the more important. By moving beyond black-box benchmarks to analyze internal representations, our study highlights how the geometry of the residual stream representations of ICL prompts can be used to both explain and anticipate the failure of foundation models to generalize.

\section{Related Work}
\label{sec:Related Works}

\paragraph{Understanding In-context Learning.}
The advent of ICL has revealed not only the potential for a widely applicable ML system stemming from a single training regiment \citep{brown2020} but also the difficulty of understanding how the model achieves this capability \citep{olsson2022context,elhage2021mathematical}. Researchers have focused on in-context regression (fitting a transformer to linear synthetic data; see Section~\ref{sec:setup}) as a model organism for ICL \citep{akyurek2023, garg2023, goddard2024specializationgeneralization, raventos2023pretraining, shen2023pretrained}.
Due to the simple nature of linear regression, a rich empirical and theoretical understanding of its traditional solutions can be applied to develop hypotheses about how transformers solve the task via ICL.

\citet{garg2023} demonstrated that transformers trained to perform in-context regression could simulate a number of different regression models, even when a modest distribution shift was induced between the in-context examples and the final query example. \citet{akyurek2023} showed that gradient descent in a linear regression setting could in principle be implemented using a simplified transformer architecture, 
a necessary theoretical underpinning for the idea that transformers could learn to implement known regression rules. Finally, \citet{von2023transformers} provided empirical evidence that transformers without MLP layers do learn a set of weights similar to the handcrafted ones proposed by \citet{akyurek2023}.

Our work challenges the hypothesis that transformers implement known regression rules when trained on in-context regression. Our argument will rely heavily on the trained transformer's dependence on its training distribution, and how asymmetries between training and testing performance cannot be accounted for using typical regression rules. Other works did not present such an example; \citet{garg2023} presented an OOD generalization experiment that succeeded, but likely due to the fact that the distribution shift occurred \textit{in-context}, rather than between the \textit{pre-training and query distributions}. 

Perhaps most similar to our work is that of \citet{goddard2024specializationgeneralization}, which also evaluates the OOD ICL generalization of transformers pre-trained on linear regression data. Our approach differs through our use of subspace distribution shifts during evaluation---going beyond the geometrical span of the pre-training data---to induce a larger generalization gap, which underscores the importance of the pre-training data in ICL capabilities and how these capabilities differ from classic statistical methods.

\paragraph{Realizing Regression Through Alternate Learning Objectives.} A notable recurring theme within machine learning research is that a single predictor can often be derived through multiple distinct training procedures.\footnote{The concept of model multiplicity~\citep{marx2020predictive} in overparameterized models highlights the converse problem: that the same dataset and training procedure can lead to multiple distinct models.}
As a rudimentary example, consider that gradient-based updates to a linear regression model will converge to the closed-form OLS solution under certain conditions~\citep{goldstein1962cauchy,bach2024learning}.
A more interesting observation is that gradient-based linear regression training with additive Gaussian input noise can be functionally equivalent to the closed-form ridge regression solution~\citep{bishop1995}. 
Works along these lines, which seek to theoretically characterize how two learning algorithms arrive at the same solution, help to motivate our work and prior related work investigating the extent to which pre-trained transformers match the performance of classic statistical methods like OLS and ridge regression.
A distinct but complementary line of work examines how regression solutions differ based on the number of feature dimensions and training samples; recent work along this direction has identified and characterized a ``double-descent'' behavior, where test risk acts non-monotonically as a function of number of samples for underdetermined linear regression trained on noisy data \citep{doub-desc}.

\begin{figure*}
    \centering    
    \includegraphics[width=0.9\linewidth]{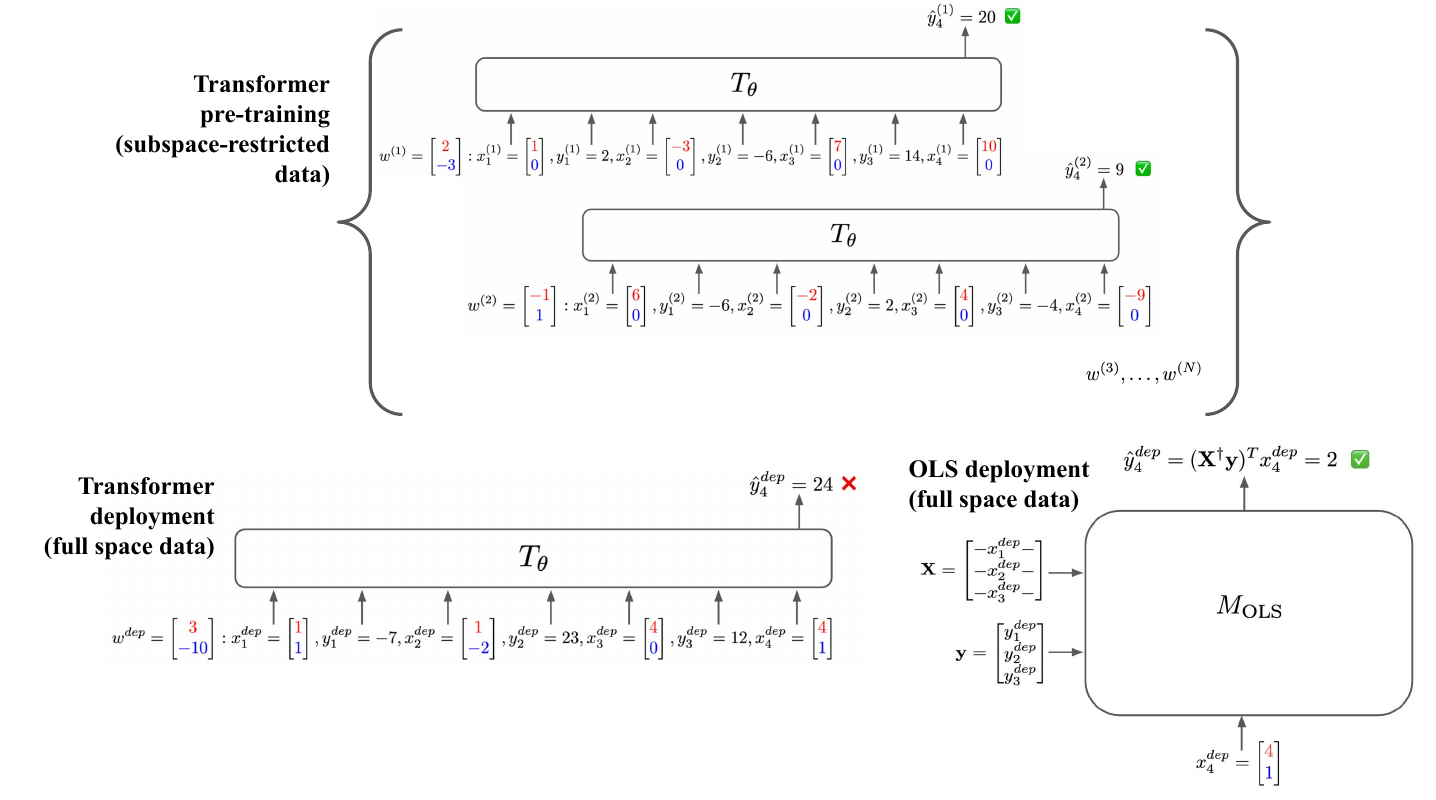}
    \caption{
    To examine the generalization of in-context learning in transformers, we pre-train on synthetic regression data with inputs drawn from a restricted subspace (i.e. the training subspace {\color{blue}$D_{\parallel}$}), then evaluate on the full subspace (i.e. $D_\square = {\color{blue}D_\parallel} \oplus {\color{red}D_\perp}$, where ${\color{red}D_\perp}$ is the orthogonal subspace).
    Unlike Ordinary Least Squares (OLS) regression, the transformer fails to generalize when deployed on $D_\square$, which includes components outside of its pre-training data.
    }
    \label{fig:fig1}
\end{figure*}

\section{Setup}
\label{sec:setup}

In this section, we describe our training and evaluation procedures. We follow the framework established by \citet{garg2023, kotha2024} to generate in-context linear regression prompts.
A GPT-2-style transformer is then pre-trained over a collection of such prompts with different ground truth weight vectors, causing the model to learn an ICL solution.
We then compare the transformer’s performance to that of classical regression techniques on ICL prompts sampled using held-out ground truth weight vectors.

\subsection{Data Generation and Transformer Training Setup}
\label{sec:training_setup}
We train a transformer model $T_\theta$ on a collection of in-context learning (ICL) prompts for linear regression. Each prompt is drawn from a prompt distribution denoted by \( \mathcal{D}(P_w, P_x), \) with the generation process as follows:

\begin{enumerate}
    \item \textbf{Task Generation:}  
    A weight vector is drawn as 
    \[
    w:= P_ww_g, \quad w_g \sim \mathcal{N}(0, \mathbf{I}_d).
    \]
    Weights are projected using a projection $P_w$. We also refer to this final weight vector $w$ and the regression problem it induces as a \textit{task}. The procedure for creating the projection is outlined in Appendix~\ref{sec:proj_matrix} .
    
    \item \textbf{Data Sampling:}  
    For each prompt, we independently sample $k+1$ input-target\footnote{In our implementation, we append 19 zeros to $y_i$ to match the token dimensionality.} pairs of the form:
    \[
    x:= P_xx_g, \quad x_g \sim \mathcal{N}(0, \mathbf{I}_d),\quad     y := w^\top x.
    \]
    Inputs are projected by an orthogonal projection $P_x$ .

    We also consider the impact of adding label noise in this regime. The results of these experiments are presented in Appendix~\ref{sec:label_noise}.

    \item \textbf{Prompt Creation:}      
    A prompt comprises $k$ labeled in-context exemplars followed by a single query point:
    \[
    p = \bigl(x_1, y_1, x_2, y_2, \ldots, x_k, y_k, x_{k+1}\bigr).
    \]
\end{enumerate}

$T_\theta$ predicts each token conditioned on all previous tokens. For ease of reference, we make this conditioning on the tokens preceding an input $x_i$ in a prompt implicit in our notation. We do the same thing for generic models $M$ that are trained on the corresponding context, such as OLS:

\[
T_\theta(x_i) := T_\theta(x_i|(x_1,y_1,...,x_{i-1}, y_{i-1})),
\]
\[
M(x_i) := M(x_i|(x_1,y_1,...,x_{i-1}, y_{i-1})).
\]

For our experiments, we fix the dimensionality at $d=20$ and set $k=40$. Transformers are randomly initialized and then trained on a large corpus of prompts  
from the specified distribution \( \mathcal{D}(P_w, P_x) \). These pre-trained transformers are then used to do in-context learning on new prompts coming from varying distributions 

\subsection{Regression Baselines}

In our experiments we compare against:

\paragraph{Ordinary Least Squares (OLS):}  
We compute the OLS solution using the pseudoinverse:
\begin{align}
    &\hat{\boldsymbol{\beta}}_{OLS} = \mathbf{X}^\dagger \mathbf{y} = (\mathbf{X}^\top \mathbf{X})^{-1} \mathbf{X}^\top \mathbf{y}, \nonumber \\
    &M_{OLS}(x_i) := \hat{\boldsymbol{\beta}}_{OLS}^\top x_i \nonumber.
\end{align}
To emulate the transformer’s auto-regressive behavior, when predicting $x_{i+1}$ we compute $\hat{\boldsymbol{\beta}}_{OLS}$ using the first $i$ examples of the corresponding prompt,
i.e. $\mathbf{X} = [x_1, \ldots x_k]^T$ and $\mathbf{y} = [y_1, \ldots y_k]^T$.

\paragraph{Ridge Regression:}  
The ridge regression solution is given by:
\begin{align}
    &\hat{\boldsymbol{\beta}}_{Ridge} = (\mathbf{X}^\top \mathbf{X} + \lambda \mathbf{I}_d)^{-1} \mathbf{X}^\top \mathbf{y}, \nonumber \\
    &M_{Ridge}(x_i)  := \hat{\boldsymbol{\beta}}_{Ridge}^\top x_i. \nonumber
\end{align}
where $\lambda$ is the regularization parameter. This is similarly trained in an auto-regressive manner.

We also compare the transformer's behaviour with a Bayesian regression method as well as a regressor trained with gradient descent. A thorough description of those additional methods, evaluation setup and training can be found in Appendix~\ref{sec:baselines}

\section{Transformers Don’t In-Context Learn Least Squares}
\label{sec:transformers_dont}

In this section, we present a series of experiments designed to probe how restricting the diversity of training tasks or training inputs affects the learned regression behavior. Our experiments compare transformer performance with several classical regression baselines in OOD settings.

Figure~\ref{fig:fig1} illustrates 
a simple case ($d = 2$) of 
the induced distribution shift between pre-training and evaluation.

Two complementary experimental setups are considered. In both, we consider randomly created orthogonal subspaces $A$ and $B$ of $\mathbb{R}^d $ such that $ \forall a\in A, b\in B, a^\top b = 0.$
\begin{enumerate}[label=\alph*)]
    \item \textbf{Weight Subspace Restriction:} We project the training weight vectors onto a lower-dimensional subspace. Specifically, if $P_A$ is a projection matrix to $A$, we train the transformer on prompts from the distribution $\mathcal{D}(\, P_w=P_A, \,P_x=\mathbf{I}_d)$.

    \item \textbf{Input Subspace Restriction:} We project the prompt feature vectors onto a lower-dimensional subspace. Specifically, if $P_A$ is a projection matrix to $A$, we train the transformer on prompts from the distribution $\mathcal{D}(\, P_w=\mathbf{I}_d, \,P_x=P_{A})$.
\end{enumerate}

In both cases, we elect for $A$ and $B$ to be 10 dimensional subspaces, exactly half of our chosen dimensionality of $d=20$. We detail how we sample these matrices in Appendix~\ref{sec:proj_matrix}. We vary the dimensionality of the training subspace in Appendix \ref{sec:app_vary_subspace_dim}, where we find similar results.
\subsection{Results}
\label{sec:restrict-weight}
\begin{figure*}[t]
    \centering
    \includegraphics[width=0.6\linewidth]{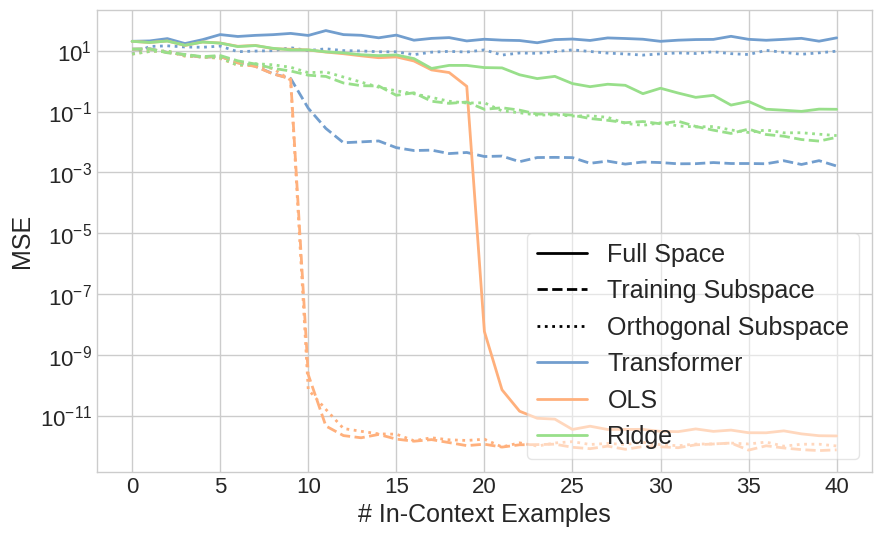}
    \hfill
    
    \caption{Prediction error of transformers trained on prompts with an input projection, shown for prompts with inputs in the training subspace, orthogonal complement, and entire space.
    Note that OLS transitions from underdetermined to overdetermined more quickly in the training/orthogonal subspace compared with the full space, as the former have half the rank of the latter.
    }
    \label{fig:input_proj_1}
\end{figure*}

We train transformers to perform in-context regression in both of these settings and present the results for input subspace restriction below. We found very similar results for weight subspace restriction, and so we defer those results to Appendix \ref{sec:app_weight_space_restriction} as the analysis and insights are identical. To simplify the notation, we introduce the shorthand $\mathcal{D}(P=P_A) := \mathcal{D}(P_w =\mathbf{I}_d,  P_x=P_A)$.

After training a transformer on prompts from the distribution $ \mathcal{D}(P=P_{A})$, which we refer to as the \textit{training subspace} $D_\parallel$, we evaluate the transformer and our regression baselines on prompts from the training subspace as well as two other distributions, which we refer to as the \textit{orthogonal subspace} $D_\perp$ and \textit{full space} $D_\square$, respectively:

\begin{equation}
    D_{\parallel} := \mathcal{D}(P=P_{A}), \quad D_{\perp}: =\mathcal{D}(P=P_{B}),
\end{equation}
\begin{equation}\nonumber
    D_{\square} :=\mathcal{D}(P=\mathbf{I}_d).
\end{equation}

$P_B$ is a projection matrix to $B$, the subspace orthogonal to the subspace $A$. We present the results of these evaluations in Figure \ref{fig:input_proj_1}.
The transformer performs well when \(w\) lies in the training subspace, similar to OLS.
However, its performance degrades heavily when 
$w$
falls in the orthogonal subspace. 

\begin{figure}[h]
    \centering
    \includegraphics[width=0.6\linewidth]{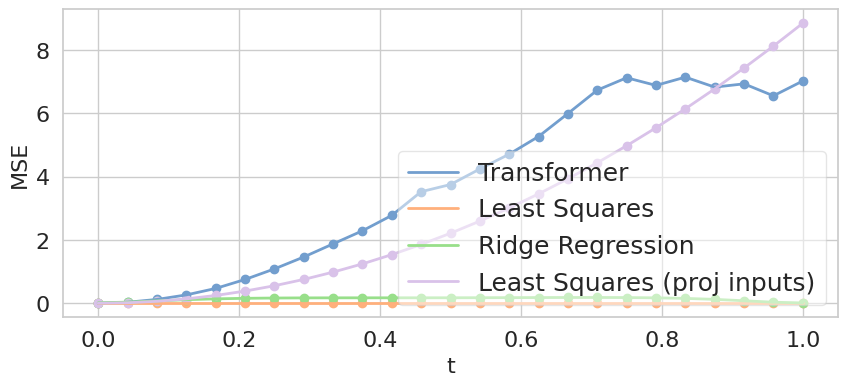}
    \hfill
    
    \caption{Prediction error as we blend an in‑distribution inputs with OOD inputs (Equation~\ref{eq:convex-combo}). The ``Least Square (proj inputs)'' baseline corresponds to running OLS on the test prompt after projecting it onto the training subspace (which incurs error because it ignores some features of the test prompt). Note the the performance discrepancy between the transformer $T_\parallel$ and the projected OLS baseline reveals that the transformer is not simply performing OLS on the components of the input within the training subspace.
    }
    \label{fig:input_proj_2}
\end{figure}

To further characterize this behavior, Figure~\ref{fig:input_proj_2} presents the error as the test example is formed by a convex combination:
\begin{align}
x_i = t(P_Ax_i) + (1-t)(P_{B}x_i),\qquad t\in[0, 1].
\label{eq:convex-combo}
\end{align}
The gradual increase in error as \(t\) decreases confirms that the transformer’s learned ICL solution is biased toward the training subspace.

In addition to the notable disparities in OOD performance between the regression baselines and the input restricted transformer, the \textit{in-distribution} performance is also vastly different between the two sets of models. Notably, for predictions corresponding to \(i < d\) (where the regression problem is underdetermined), all models exhibit similar performance. However, once we enter the overdetermined regime, the OLS baseline achieves an error that is several orders of magnitude lower than that of the transformer models or the Ridge baseline.

This result, observed under a basic in-distribution setting, clearly demonstrates \emph{transformers do not emulate the optimal least-squares solution, even in-distribution}. In other words, our initial experiments on out-of-distribution settings are underpinned by a more fundamental disparity: OLS consistently outperforms transformer-based in-context regression in-distribution. This challenges previous claims that transformers can inherently learn regression rules on par with OLS.

\section{Identifying ICL Generalization Through Spectral Signatures}
\label{sec:spec_sig}
\subsection{Measuring Distribution Shift in the Transformer Residual Stream}

So far we have shown that successful ICL generalization in transformers depends implicitly on \emph{where} the prompt lies relative to the pre-training data.
Next we examine \emph{why} such sensitivities to distribution shift occur.
Our strategy is to trace the internal representations that the transformer assigns to each token and compare those representations between test prompts from $D_\parallel, D_\perp$, and $D_\square$.

In this analysis, we consider the two transformer models: one transformer $T_\parallel$ trained on input-restricted prompts from $D_\parallel$, and another $T_\square$ trained on unrestricted prompts from $D_\square$.

\paragraph{Collecting representations.}
Because the transformer operates auto-regressively, it processes the prompt $p = (x_1, y_1, x_2, y_2, \ldots, x_k, y_k, x_{k+1})$ one token at a time. As each token $t_i \in p$ passes through the network, it acquires its own residual‑stream representation and produces a corresponding prediction. Let $z(t_i)\in\mathbb{R}^m$ denote the residual‑stream vector of $t_i$ immediately before the final readout head.

We collect the representations for the tokens in a prompt 
into a matrix and denote it as:

  \begin{align}
    Z_p &= \begin{bmatrix}
           z(x_1) \\
           z(x_2) \\
           \vdots \\
           z(x_{k+1})
         \end{bmatrix} \in\mathbb{R}^{ (k+1)\times d}
  \end{align}

For every prompt distribution, we collect a batch of $b = 128$ prompts to create:
\begin{equation}\label{eqn:distributions}
    S_\parallel = \{Z_{p_i}; p_i \sim D_{\parallel}\}_{i=1}^b, \quad S_\perp = \{Z_{p_i}; p_i \sim D_{\perp}\}_{i=1}^b,
\end{equation}
\begin{equation}\nonumber
    S_\square = \{Z_{p_i}; p_i \sim D_{\square}\}_{i=1}^b.
\end{equation}
Finally, for every representation matrix $Z_p$ we can compute its singular value decomposition (SVD) as $Z_p = U_p\Sigma_p V^\top_p$.

\begin{figure}[t]
  \centering
  \subfigure[Singular-value spectrum of representations for the input-restricted transformer $T_\parallel$.]{
    \includegraphics[width=0.42\textwidth]
    {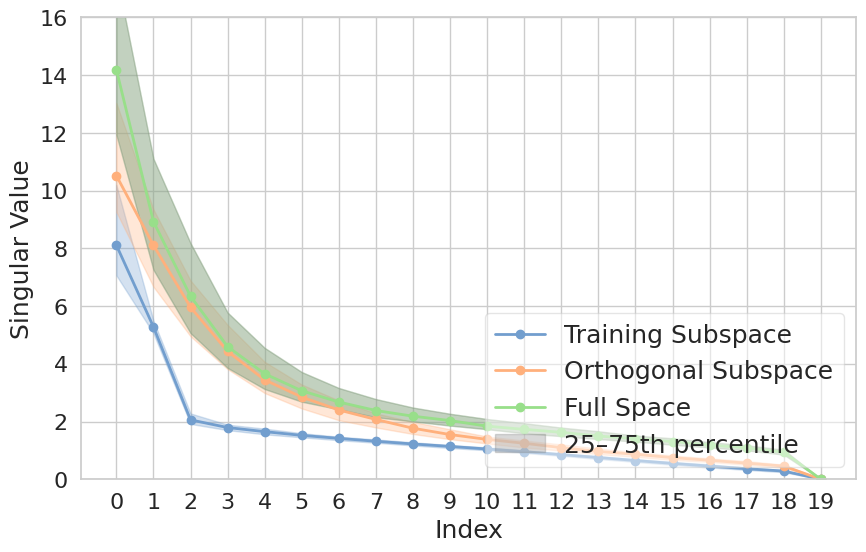}
    \label{fig:eigenspectrum_input_1}
  }
  \hspace{0.05\textwidth}
  \subfigure[Singular-value spectrum of representations for the baseline transformer $T_\square$.]{
    \includegraphics[width=0.42\textwidth]{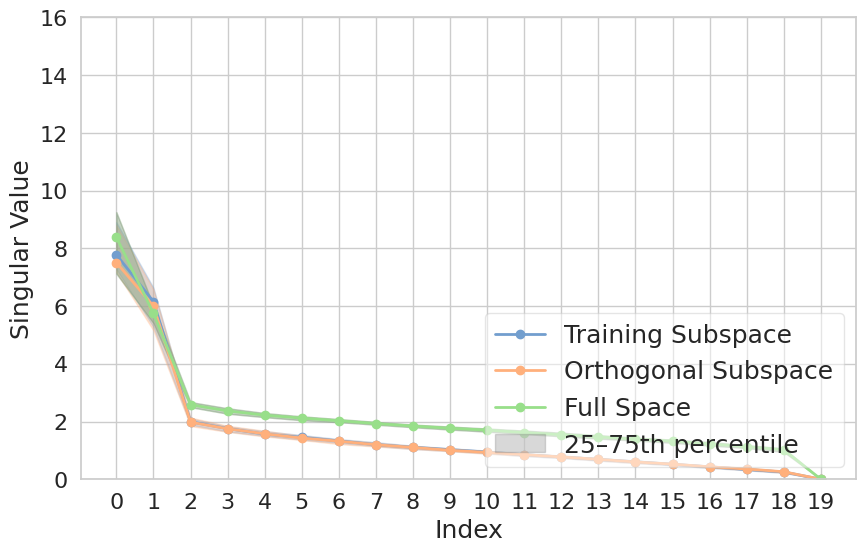}
    \label{fig:eigenspectrum_baseline_1}
  }

    \caption{Singular‑value spectra of residual‑stream embeddings for \textit{training subspace, orthogonal subspace} and \textit{full space} prompts.}
    \label{fig:eigenspectrum_1}
\end{figure}

\paragraph{Spectral distribution. }

Figure~\ref{fig:eigenspectrum_input_1} plots the singular‑value spectra for each prompt distribution described in Equation~\ref{eqn:distributions} for the transformer $T_\parallel$.
The spectra corresponding to $S_\parallel$ exhibits a steep drop, and the leading 2 singular values dominate, in addition to the low variance of the singular values.  
OOD prompts lack this structure, their spectra are flatter and have a much higher variance across the batches, and higher values in general.

\subsection{Spectral Signatures Vary by Distribution}

To measure the stability of the directions of maximal variance, we choose a `canonical' set of right singular vectors for each distribution listed in Equation \ref{eqn:distributions} by taking a batch of 20 prompts, pooling their representations into a matrix $Z^*$, and extracting the right singular vector matrix $V^{*\top}$.

\begin{figure*}[htb]
  \centering
  \subfigure[Cosine similarity between singular vectors of residual representations from model $T_\parallel$ and its `canonical' set of singular vectors.]{%
    \includegraphics[width=0.45\textwidth]{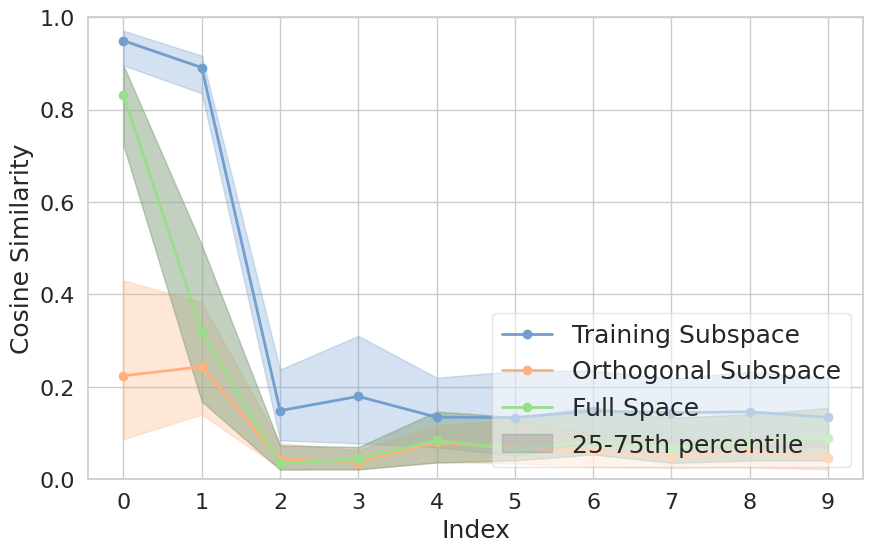}
    \label{fig:eigenspectrum_input_2}
  }
  \hspace{0.05\textwidth}
  \subfigure[Cosine similarity between singular vectors of residual representations from model $T_\square$ and its canonical set of singular vectors.]{%
    \includegraphics[width=0.45\textwidth]{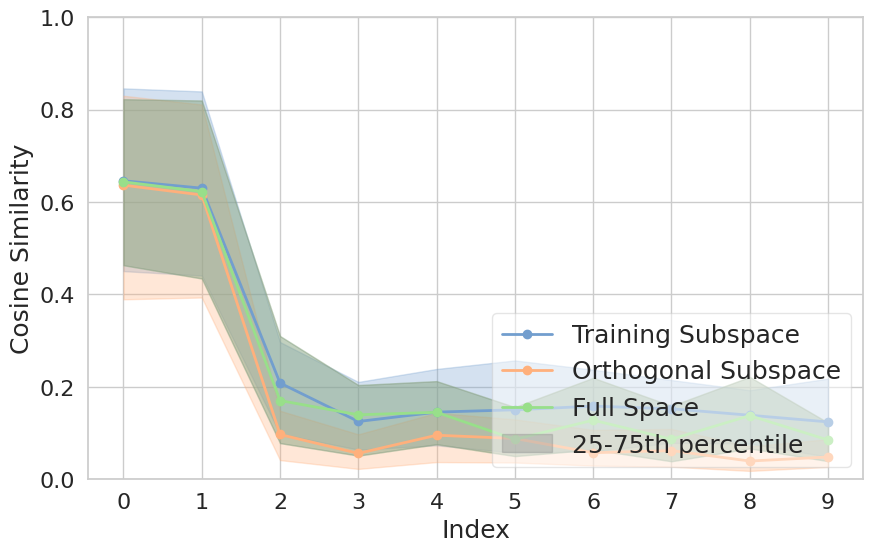}
    \label{fig:eigenspectrum_baseline_2}
  }

    \caption{Alignment of singular‐vector directions across prompts. For each distribution, we compute cosine similarities between each prompt’s singular vectors and a fixed reference (`canonical') set, showing that in‐distribution prompts yield highly consistent first two components, whereas OOD prompts exhibit greater variability.}
    \label{fig:eigenspectrum_2}
\end{figure*}

We then compute cosine similarities index-wise of the right singular vectors for every representation matrix $Z_p$ with the corresponding canonical singular vectors from $p$'s distribution:

\begin{equation}\label{eqn:signature_vector}
    C_p := diag(V^{*\top}V_p).
\end{equation}

Mean values for the indices of $C_p$ are shown in Figure~\ref{fig:eigenspectrum_input_2} for $T_\parallel$.
For $D_{\parallel}$'s representations from $T_\parallel$, the restricted transformer, the first two singular vectors are almost identical across prompts, indicating that the representation for these prompts present the same pattern of high variance along two fixed directions.  
For $D_{\square}$ and $D_{\perp}$ the principal directions vary, reinforcing the view that the model has no reliable final representation for those inputs. We denote the presence of these two fixed principal directions as a \textit{spectral signature}, as they distinguish in-distribution and OOD prompts.

We now consider the baseline transformer $T_\square$. As shown in Figure~\ref{fig:eigenspectrum_baseline_1} and Figure~\ref{fig:eigenspectrum_baseline_2}, $T_\square$ demonstrates this spectral signature in the representation for prompts both in \textit{and} outside of $D_\parallel$, as it was trained on $D_\square$. We demonstrate the presence of this signature in other training regimes in Appendix~\ref{sec:app_weight_space_restriction}.

Finally, we demonstrate that the distribution of the first two values of $C_p$ can reliably differentiate in-distribution representations from OOD representations. Specifically, we calculate $C_p$ on a subset of the prompts from $S_\parallel$ and fit a bivariate Gaussian to the first two indices of each $C_p$. We then determine whether hold-out projections $C_p$ from prompts either from $D_\parallel$ or $D_\perp$ lie in the 95\% confidence region defined by that Gaussian. Further details are provided in Appendix \ref{sec:app_ood_detect}.

Results are presented in Table \ref{tab:ood_detector}, where we see that representations derived from prompts from the training distribution are within the 95\% confidence region of the Gaussian far more than representations from $S_\perp$. This further demonstrates how the spectral signature can characterize the distribution data is sampled from.

\begin{table}[t]
    \centering
    \begin{tabular}{|c|c|c|}
        \hline
        $.$ & $S_\parallel$ & $S_\perp$\\ 
        \hline
        $T_\parallel$  &   $94.45\% \pm 4.29\%$   & $19.39\% \pm 12.46\%$          \\
        \hline
        $T_\square$ & $91.48\% \pm 3.63\%$ & $92.9\% \pm 1.54\%$\\
        \hline
    \end{tabular}
    \caption{Mean $\pm$ standard deviation of the \% of prompts within the 95\% confidence region defined by a Gaussian fit to the $C_p$ projection vectors of representations from $S_\parallel$.}
    \label{tab:ood_detector}
\end{table}

\subsection{Spectral Signatures Impact Prediction}
We next probe at how the fixed representation subspace defined by the two fixed `canonical' singular vectors affects prediction for the transformers $T_\parallel$ and $T_\square$.
Taking $f$ as the transformer's readout head we compare $Zf$ and $P_i Zf $ where 
 $P_i$ is the orthogonal projection onto the subspace defined by the respective model's  first two canonical singular vectors.

Table~\ref{tab:representation_norms} shows:
\begin{enumerate}[label=\alph*)]
\item \textbf{Canonical subspace dominance:} For both $T_\parallel$ and $T_\square$,  most of the final‐layer representation’s norm lies in the span of their respective canonical singular vectors.
\item \textbf{Prediction preservation:} As a result, projecting onto either the respective canonical subspace preserves nearly all of the model’s prediction.
\end{enumerate}

\begin{table}[t]
    \centering
    \begin{tabular}{|c|c|c|}
        \hline
        $.$ & $\text{$\frac{\|P_iZ\|}{\|Z\|}$}$ & $\text{$MSE(Zf, P_iZf)$}$ \\ 
        \hline
        $T_\parallel$  &   0.985    & 0.195            \\
        \hline
        $T_\square$ & 0.97 & 0.09\\
        \hline
    \end{tabular}
    \caption{Magnitude of the projection of $Z$ onto the top two canonical subspace and MSE between predictions from projected and original embeddings.
    }
    \label{tab:representation_norms}
\end{table}

\begin{table}[t]
    \centering
    \begin{tabular}{|c|c|c|}
        \hline
        $.$ & $\frac{||(V^\top_pf)_{:2}||^2_2}{||f||^2_2}$ & $\frac{||(V^\top_pf)_{2:10}||^2_2}{||f||^2_2}$\\ 
        \hline
        $S_\parallel$  &    $0.897 \pm 0.012$   & $0.008 \pm 0.005$          \\
        \hline
        $S_\perp$ & $0.352 \pm 0.205$ & $0.388 \pm 0.182$\\
        \hline
    \end{tabular}
    \caption{Mean $\pm$ standard deviation of normalized $T_\parallel$ readout head projections.}
    \label{tab:weight_vec_dot_prods}
\end{table}

In Table \ref{tab:weight_vec_dot_prods}, we plot average normalized squared dot products of the singular vectors of representation matrices from $S_\parallel$ and $S_\perp$ with the readout head of the transformer $T_\parallel$. Specifically, for readout head $f$, we calculate $V^\top_pf$, and index into its first two coordinates as $(V^\top_pf)_{:2}$ and the third through tenth coordinates as $(V^\top_pf)_{2:10}$. We see that for the representations coming from $S_\parallel$, the first two vectors tend to constitute around 90\% of the magnitude of the readout vector, while the next eight singular vectors constitute less than 1 percent on average. Meanwhile, for $S_\perp$, $(V_p^\top f)_{:2}$ and $(V_p^\top f)_{2:10}$ both constitute just over a third of the total magnitude of $f$. This demonstrates that the dependency of predictions on the spectral signature is not determined simply because the majority of representational variance is along the first two singular vectors.

\begin{figure}[h!]
    \centering
    \includegraphics[width=\linewidth]{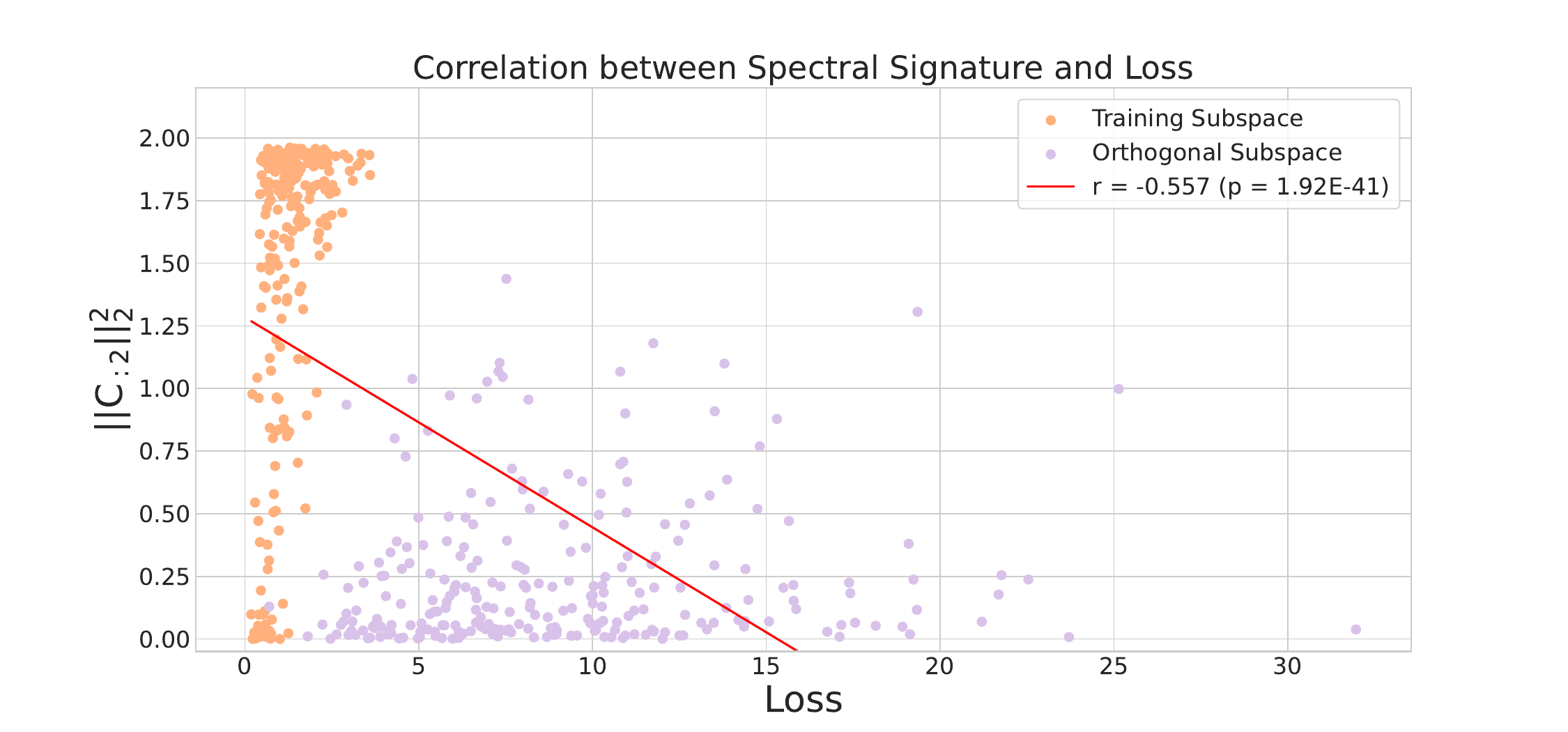}
    \hfill
    \caption{Scatter plot of $||C_{p, :2}||^2_2$ vs MSE over all token positions for prompts from $D_\parallel$ and $D_\perp$. Line of best fit is depicted along with associated Pearson correlation.}
    \label{fig:corr_loss_w_sig}
\end{figure}

Additionally, we demonstrate a relationship between the vector $C_p$ from Equation \ref{eqn:signature_vector} and the average loss of the transformer on a given prompt. Specifically, if we denote the first two indices of $C_p$ as $C_{p, :2}$, we will correlate $||C_{p, :2}||^2_2$ with the average mean squared error across token positions. We see in Figure \ref{fig:corr_loss_w_sig} that there is a clear, statistically significant relationship between $||C_{p, :2}||^2_2$ and the average loss when considering prompts from both $D_\parallel$ and $D_\perp$.

\section{Conclusion}
\label{sec:conclusion}

\paragraph{Limitations.}
Our experiments focus on synthetic linear‐regression prompts and GPT‑2–style architectures; the extent to which these findings generalize to richer, real‐world ICL tasks (e.g.\ language‐translation, code generation) remains an open question. We provide a preliminary analysis of spectral signatures in pre-trained LLMs in Appendix~\ref{sec:app_LLM}. Moreover, while our spectral signature serves as a useful diagnostic and OOD detector, we have not yet fully characterized how it emerges during pretraining or how it interacts with specific architectural components (e.g.\ layer normalization, attention heads, MLP layers).

\paragraph{Discussion.}

We have demonstrated that transformers’ in-context regression abilities hinge on a tight coupling between pretraining data and representation geometry, rather than on an emulation of known learning rules like OLS. Our out-of-distribution experiments reveal that small, but structured, shifts in input or weight subspaces can cause large degradations in performance. This is a critical concern for any foundation model deployed in complex, real-world environments.

By uncovering a \emph{spectral signature} for this task, stable variance within a low-dimensional subspace and structured singular value spectra, we identify a promising indicator of when a transformer may be operating outside its training distribution. While further work is needed to establish robust thresholds and characterize the spectral signature of other tasks, this geometric cue could inform real-time monitoring or trigger human review when a model encounters atypical prompts.

We see several promising avenues for future work regarding this formulation and the spectral signature:
\begin{enumerate}
    \item \textbf{Training Dynamics} of the spectral signature to track how it evolves and stabilizes over training. This could involve examining gradient-updates and how they relate to the residual-stream representation of the in-distribution and OOD subspaces. 
    \item \textbf{Empirical validation} of spectral-signature thresholds on richer, real-world ICL tasks (e.g.\ language, multimodal settings).
    \item \textbf{Integration} of this indicator into monitoring pipelines, exploring how it combines with other reliability signals to guide fallback strategies.
\end{enumerate}

By moving beyond benchmarks to the underlying mechanisms of failure and identifying concrete metrics for further study, our work lays groundwork for more reliable and responsible deployment of foundation models.

\section*{Acknowledgements}
The resources used in preparing this research were provided, in part, by the Province of Ontario, the Government of Canada through CIFAR, and companies sponsoring the Vector Institute~\url{www.vectorinstitute.ai/partnerships/}. 
Joshua Hill was supported in part by an NSERC Undergraduate Student Research Award.

\bibliography{references}
\bibliographystyle{plainnat}

\newpage
\appendix
\section{Projection Matrix Details}
\label{sec:proj_matrix}
To create the projection matrix $P$, we first take a $(d, q)$ random Gaussian matrix $V$, and compute $Q, U = QR(V)$. We then take the first $q$ columns of $Q$, and compute $P$ as the matrix product of these columns multiplied with itself. The projection matrix $P_{orth}$ to the orthogonal space is computed as $P_{orth} = I - P$.

\section{Adding Label Noise}
\label{sec:label_noise}
To complement the original experiments, we train a transformer $T_\square^\star$ on prompts from the distribution $\mathcal{D}(P_w = \textbf{I}_d,\, P_x = \textbf{I}_d)$ where the prompt's labels are replaced by:
    \[
    y_i = w^\top x_i + \epsilon,\quad \epsilon \sim \mathcal{N}(0, \sigma^2 \mathbf{I}_d).
    \]

\begin{figure}[h]
    \centering
    \includegraphics[width=0.6\textwidth]{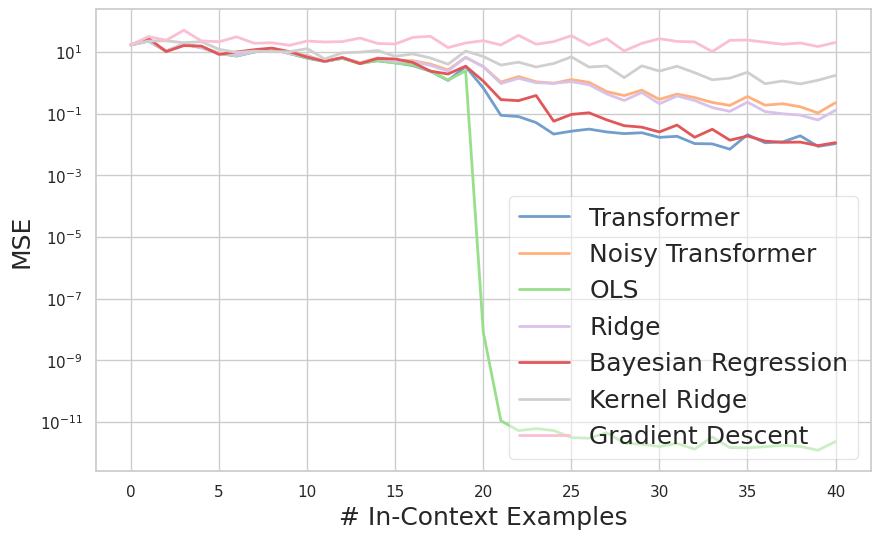}
    \hfill
    \caption{Prediction error of several models (Section \ref{sec:baselines}), including transformers trained on prompts with and without noisy labels and tested on prompts without label noise.}
    \label{fig:noisy_1}
\end{figure}

We evaluate models on $D_\square$ \textit{which does not have noise added to the labels}.
In Figure~\ref{fig:noisy_1} we plot the performance of several models, most notably transformers trained with and without noisy labels ($T_\square$ and $T_\square^\star$, respectively). We observe that in the noiseless test set, $T_\square^\star$ achieves performance similar to ridge regression and poorer than OLS and $T_\square$. This result further demonstrates how the learned mechanism, and therefore generalization, of transformers trained on this task is closely linked to the pre-training distribution of the model.

\begin{figure}[h]
    \centering
    \includegraphics[width=0.6\textwidth]{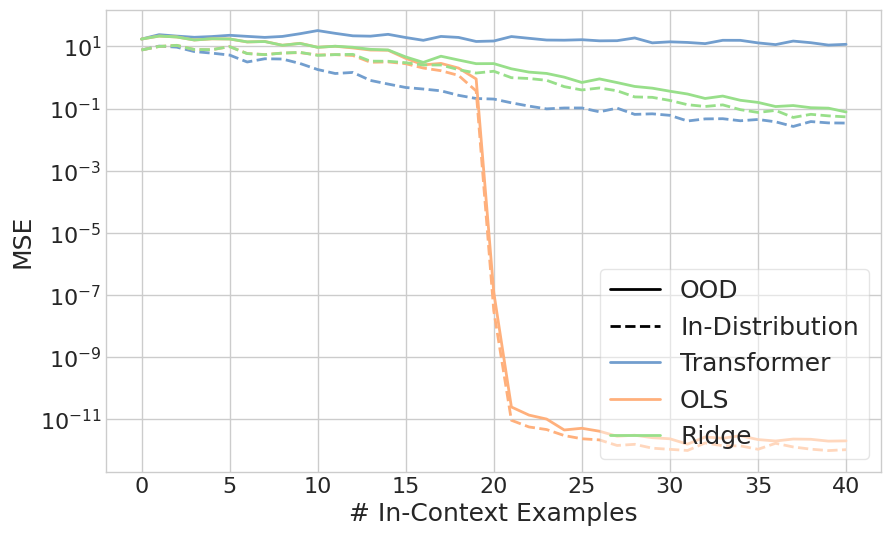}
    \hfill
    \caption{Prediction error of a transformer trained on a noisy variant of $D_\parallel$ and tested on noiseless prompts from $D_\perp$.}
    \label{fig:noisy_ood}
\end{figure}

We also experiment with a transformer trained on a variant of the distribution $D_{\parallel}$ with added label noise, and test it on prompts from $D_\perp$. In Figure \ref{fig:noisy_ood}, we see that this model also struggles with this distribution shift.

\section{Experimental Details}
\label{sec:baselines}

In our experiments we compare against:

\paragraph{Ordinary Least Squares (OLS):}  
We compute the OLS solution using the pseudoinverse:
\[
\hat{\boldsymbol{\beta}}_{OLS} = \mathbf{X}^\dagger \mathbf{y} = (\mathbf{X}^\top \mathbf{X})^{-1} \mathbf{X}^\top \mathbf{y}.
\]
To emulate the transformer’s auto-regressive behavior, when predicting $x_{i+1}$ we compute $\hat{\boldsymbol{\beta}}_{OLS}$ using the first $i$ examples.

Note in Section~\ref{sec:transformers_dont}, we make use of a projected input OLS baseline, where each $x_i$ is orthogonally projected into the trainign subspace, in order to provide evidence against the claim $T_\parallel$ is performing OLS on the components of $x_i$ in the training subspace (with the corresponding non-projected labels). 

\paragraph{Ridge Regression:}  
The ridge regression solution is given by:
\[
\hat{\boldsymbol{\beta}}_{Ridge} = (\mathbf{X}^\top \mathbf{X} + \lambda \mathbf{I}_d)^{-1} \mathbf{X}^\top \mathbf{y},
\]
where $\lambda$ is the regularization parameter. This is similarly updated in an auto-regressive manner.

\paragraph{Bayesian Linear Regression:}  
Bayes' rule, given by \(p(\beta \mid X, y) \propto p(\beta)p(y \mid X, \beta)\), updates a prior distribution using a likelihood distribution to produce a posterior distribution, capturing the process of updating a prior 'belief' when given new evidence. For linear regression, we encode a belief about the model parameters $\beta$, and update it according to the context examples $X, y$. Assuming a Gaussian prior on the weights,
\[
p(\beta) = \mathcal{N}(0, \tau^2 \mathbf{I}_d),
\]
where we set $\tau = 1$ to match the base distribution, and a Gaussian likelihood,
\[
p(y \mid X, \beta) = \mathcal{N}(X\beta, \sigma^2 \mathbf{I}),
\]
Bayes' rule yields the posterior distribution over $\beta$, which is also Gaussian:
\[
p(\beta \mid X, y) = \mathcal{N}(\hat{\beta}_{\text{posterior}}, \Sigma_{\text{posterior}}).
\]
With:
\[
\Sigma_{\text{posterior}} = \left(\frac{1}{\sigma^2}\mathbf{X}^\top \mathbf{X} + \frac{1}{\tau^2}\mathbf{I}_d\right)^{-1},
\]
\[
\hat{\beta}_{\text{posterior}} = \Sigma_{\text{posterior}} \left(\frac{1}{\sigma^2}\mathbf{X}^\top \mathbf{y}\right).
\]
In our experiments, the prediction for a new input $x_{i+1}$ is obtained by averaging the predictions made by $m$ weight vectors drawn from the posterior distribution.

\paragraph{Kernel Ridge Regression:}  
We also compare against kernel ridge regression, which extends ridge regression via the Representer Theorem \cite{bach2024learning}. Given a context of \(k\) in‐context examples \(\{(x_i,y_i)\}_{i=1}^k\), define the Gram matrix
\[
K \in \mathbb{R}^{k\times k}, \quad
K_{ij} = \kappa(x_i, x_j),
\]
for a positive‐definite kernel \(\kappa\), and let
\[
\mathbf{y} = \begin{bmatrix} y_1 \\ \vdots \\ y_k \end{bmatrix}.
\]
The dual coefficients \(\boldsymbol{\alpha}\in\mathbb{R}^k\) are given in closed form by
\[
\boldsymbol{\alpha}
= (K + \lambda I_k)^{-1}\,\mathbf{y},
\]
where \(\lambda > 0\) is the regularization parameter.  A prediction at a new query point \(x_{k+1}\) is then
\[
\hat{y}_{k+1}
= \sum_{i=1}^k \alpha_i\,\kappa(x_i, x_{k+1})
= \mathbf{k}(x_{k+1})^\top (K + \lambda I_k)^{-1}\,\mathbf{y},
\]
with
\[
\mathbf{k}(x_{k+1})
= \begin{bmatrix}
\kappa(x_1, x_{k+1}) \\
\vdots \\
\kappa(x_k, x_{k+1})
\end{bmatrix}.
\]
In our experiments, we use the Gaussian (RBF) kernel
\[
\kappa(x, x') = \exp\!\Bigl(-\tfrac{\|x - x'\|^2}{2\sigma^2}\Bigr).
\]

\paragraph{Gradient Descent:}  
In this baseline, we iteratively update the weight vector using gradient descent. Starting with a random normally distributed initial weight $\beta^{(0)}$, the update rule for each iteration $t$ is:
\[
\beta^{(t+1)} = \beta^{(t)} - \eta \nabla_{\beta} L(\beta^{(t)}),
\]
where $\eta$ is the learning rate and the loss is given by
\[
L(\beta) = \| \mathbf{X}\beta - \mathbf{y} \|^2.
\]
In an autoregressive setting, we initialize and update $\beta$ using the first $i$ examples to predict $x_{i+1}$.

\paragraph{Evaluation:} We assess a model $M$'s performance by evaluating predictions on a prompt $p = (x_1, y_1,...,x_k,y_k, x_{k+1})$ with associated weight vector $w_p$. We define the squared error as:
\[
\texttt{SquaredError}(M, p)_i
= \bigl(M(x_i) - w_p^\top x_i\bigr)^2 \ \forall i \in [1, k+1].
\]

When evaluating over a batch of test prompts $S_{test} = \{p_1,...,p_N\}$, we compute the mean squared error (MSE) for each index $i$:
\[
\texttt{MeanSquaredError}(M, S_{test})_i 
= \frac{1}{|S_{test}|}\sum_{p \in S_{test}} 
\texttt{SquaredError}(M, p)_i.
\]

For each $ p \in S_{test}, p \sim  \mathcal{D}(P_w, P_x)$, where the distribution $\mathcal{D}(P_w, P_x)$ varies over evaluation experiments.

\paragraph{Training: }
We use a GPT-2-style transformer with 12 layers, 8 attention heads per layer, a hidden dimension of 256, and no dropout. To yield a scalar output, we append a final linear layer that projects the token embedding down to one dimension, following \citet{garg2023}. Models are trained for 500 000 steps with the AdamW optimizer (learning rate $1\times10^{-4}$); extending to 1 000 000 or 1 500 000 steps or applying weight decay of 0 versus $1\times10^{-6}$ produced no appreciable change in performance.

Each prompt is tokenized as follows: each $x_i$ is a $20$-dimensional vector, and each $y_i$ is a $20$-dimensional vector, where the first entry contains the scalar value $y_i = w^\top x_i$ and the $19$ other dimensions are $0.$

We also adopt the curriculum schedule from \citet{garg2023}: initially the last $d_{\text{start}}=15$ dimensions of each input $x_i$ are zeroed and only $k_{\text{start}}=11$ context pairs are presented. Every 2 000 steps we decrease $d_{\text{start}}$ by 1 and increase $k_{\text{start}}$ by 2, until $d_{\text{start}}=0$ and $k_{\text{start}}=k$.

All training runs were performed on an NVIDIA RTX A4500 GPU, taking roughly 9 hours each.

\section{Scaling Prompts Experiments}
To further investigate the importance of the pre-training distribution for ICL regression, we probe how sensitive ICL is to input magnitude. We define, for each scale factor $s$, a prompt distribution 
\[
\mathcal{D}_{s} = \mathcal{D}(P_w=\textbf{I}_d,\,P_x=s\textbf{I}_d),
\]
where $s \in \{0.5, 0.75, 1.0, 1.25, 1.5, 2, 5, 10\}.$ In other words, for a prompt $p \sim \mathcal{D}_s, $ each input $x_i \sim \mathcal{N}(0, \textbf{I}_d)$ is scaled to $sx_i.$

\begin{figure}[t]
    \centering
    \includegraphics[width=0.6\textwidth]{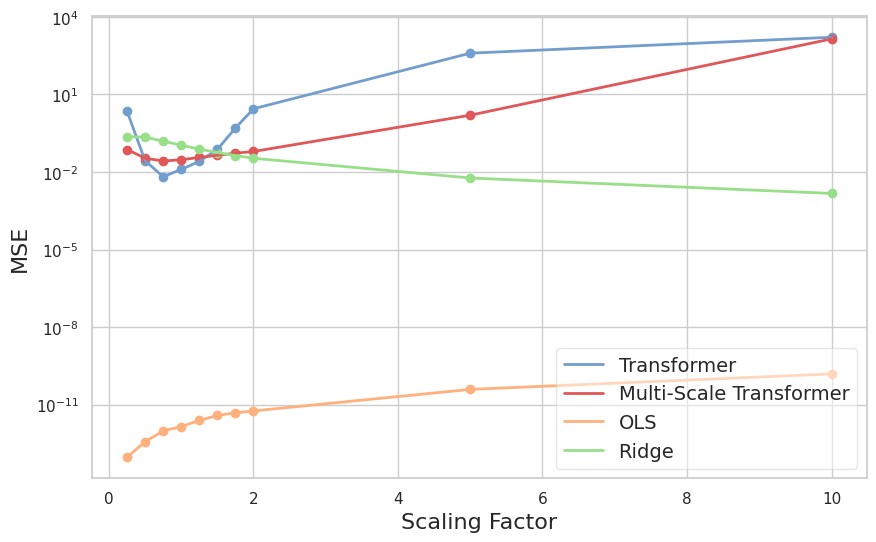}
    \hfill
    
    \caption{Comparing last‑token prediction error versus input scale.}
    \label{fig:scaling_1}
\end{figure}

Figure~\ref{fig:scaling_1} plots the final-token mean-squared error of each model as a function of $s$. The OLS and ridge baselines remain essentially flat across all $s$, whereas transformers trained on a single scale incur rapidly growing errors as  $s$ moves out-of-distribution. \citet{garg2023} ran a similar experiment where they scaled only the final token, and kept the context in-distribution, showing similar results.

We expand upon this experiment and examine the effect of training data diversity on this behavior. This is in contrast to the aforementioned experiment by \citet{garg2023}, which focused on a transformer trained without scaled task vectors. We train a new transformer
$T_\square^{\circ}$, trained on prompts drawn from:
    \[
    \mathcal{D}_{train}^\circ = \{\mathcal{D}_s : s\in\{1, 2, 3\}\}
    \]

(i.e.\ each batch uniformly samples $s \in \{1, 2, 3\}$ then draws prompts under $\mathcal{D}_s$).

Figure~\ref{fig:scaling_1} demonstrates that incorporating multiple input scales into the pre-training distribution can substantially reduce OOD degradation. This finding underscores that unlike classical regressors, transformers’ ICL performance hinges crucially on matching the scale statistics seen during training.

\section{Determining Implicit Weight Vectors}

To facilitate a more direct comparison with the regression baselines, we extract the implicitly learned weight vector $\hat{\boldsymbol{\beta}}$ from a transformer $T_\square^\circ$ that was trained on the distribution $\mathcal{D}(\, P_w=P_A, \,P_x=\mathbf{I}_d)$ as described in Section \ref{sec:app_weight_space_restriction}. Given a fixed context prompt $\hat{p}$ (with more than $d$ examples) generated by $w_j$, we predict on a set of query points $X_q$ and obtain:
\[
\hat{y}_i = T_\square^\circ(x_i; \hat{p}), \quad \forall\, x_i \in X_q.
\]
Let $\hat{\mathbf{y}}$ be the vector of predictions $\hat{y}_i$ and $\mathbf{X}_q$ be the matrix of query inputs. We then estimate the weight vector as:
\[
\hat{\boldsymbol{\beta}} = \mathbf{X}_q^{\dagger}\hat{\mathbf{y}}.
\]

\begin{table}[t]
    \centering
    \begin{tabular}{|c|c|c|}
        \hline
        $.$ & $P_A\hat{\boldsymbol{\beta}}$ & $P_B\hat{\boldsymbol{\beta}}$\\ 
        \hline
        Avg. Norm  & 0.35      & 0.04            \\
        \hline
        Variance of Norm & 0.83 & 0.04\\
        \hline
    \end{tabular}
    \caption{Average norm, variance of norm post projection of implicitly learned weight vector $\hat{\boldsymbol{\beta}}$.}
    \label{tab:1}
\end{table}

We then examine this best-fit weight vector $\hat{\boldsymbol{\beta}}$'s dependence on the subspace spanned by weight vectors used during training and those from the corresponding orthogonal subspace. Following the notation from Section \ref{sec:transformers_dont}, we specifically analyze $P_A\hat{\boldsymbol{\beta}}$ for the training subspace and $P_A\hat{\boldsymbol{\beta}}$ for the orthogonal subspace. In Table~\ref{tab:1} we calculate the average norm across multiple prompts of these two projected vectors as well as the variance of the norm. We observe that the implicit weight vector has a much greater average norm, and that the norm has greater variance, when projected into the training subspace than the orthogonal subspace. This further demonstrates that the predictive behaviour of the transformer is reliant on the pretraining distribution.

\section{Varying Training Subspace Dimension}\label{sec:app_vary_subspace_dim}
\begin{figure}[t]
    \centering
    \includegraphics[width=0.6\textwidth]{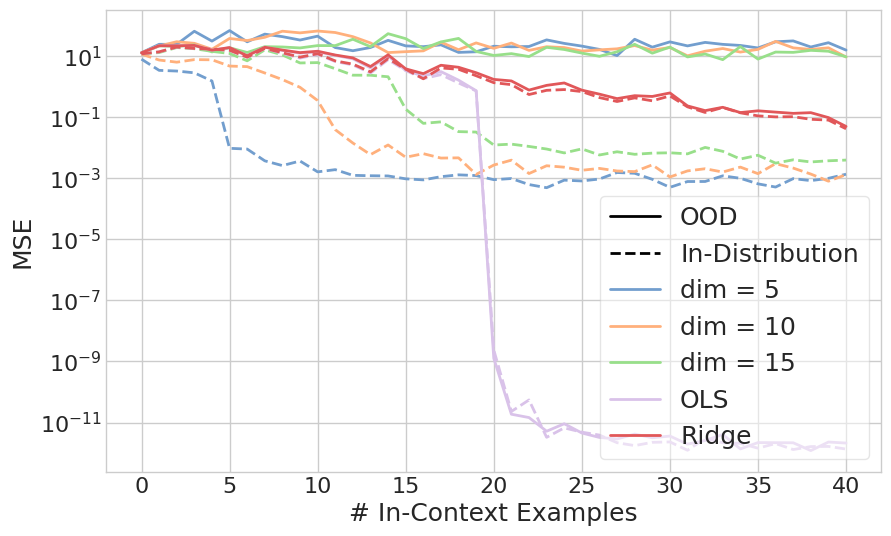}
    \hfill
    \caption{Prediction error of transformers trained on 3 different task-vector subspace dimensions, shown for in‑distribution prompts (projected into respective training subspace, full 20-dimensional space for ridge and OLS) versus out‑of‑distribution prompts (full 20‑dimensional task vector).}
    \label{fig:weight_proj_3}
\end{figure}

In Section \ref{sec:transformers_dont}, we describe a projection scheme for task vectors and inputs. Throughout the paper we primarily focus on projection into subspaces of dimension $10$. However, we can just as easily project into a subspace of any dimensionality.
To explore the effect of subspace dimensionality, we also train models with weight vector projections \(P_{A,5}\) and \(P_{A,15}\), which project into 5 and 15 dimensional subspaces, respectively. This describes a corresponding orthogonal subspace for evaluation that is 15 and 5 dimensional, respectively. Figure~\ref{fig:weight_proj_3} illustrates that the error plateaus at the dimensionality $q$ of the training subspace, consistent with the intuitive notion that the model needs to determine the coefficients of the task vector $w$ used in the prompt, which has $q$ degrees of freedom. We also observe that each of these transformers struggle to generalize out of distribution, regardless of their training subspace dimensionality.

\section{Weight Subspace Restriction}\label{sec:app_weight_space_restriction}

\begin{figure}[t]
    \centering
    \includegraphics[width=0.6\textwidth]{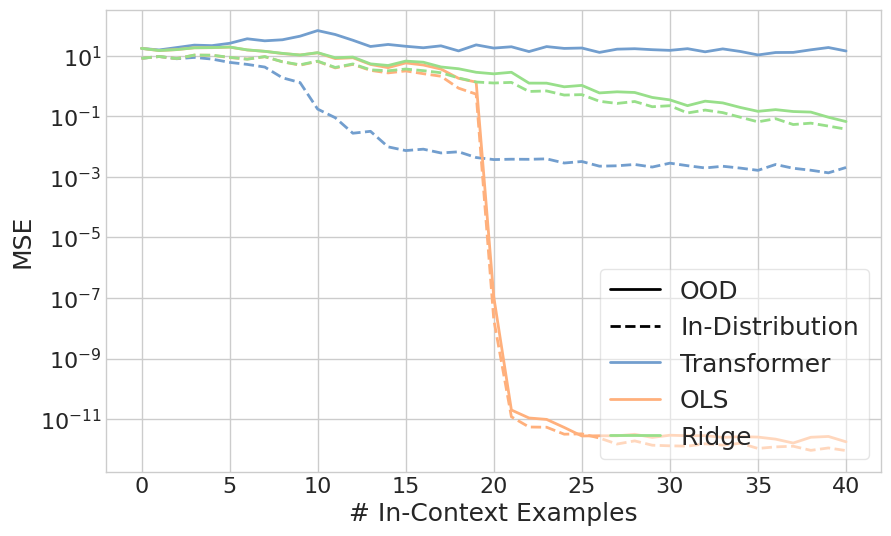}
    \hfill
    \caption{Performance on prompts from $\mathcal{D}(\, P_w=P_A, \,P_x=\mathbf{I}_d)$ (In-Distribution) and $\mathcal{D}(\, P_w=P_B, \,P_x=\mathbf{I}_d)$ (OOD) for several methods.}
    \label{fig:weight_space_id_ood}
\end{figure}

In this section, we report results for experiments analogous to those presented in Section \ref{sec:transformers_dont} with the exception that the distribution shift takes place in the \textit{weight space}, rather than in the input space . Specifically, we train a transformer on prompts from the distribution $\mathcal{D}(\, P_w=P_A, \,P_x=\mathbf{I}_d)$ and test on prompts from the distribution $\mathcal{D}(\, P_w=P_B, \,P_x=\mathbf{I}_d)$. We present performance in Figure \ref{fig:weight_space_id_ood}, where we observe nearly identical trends the results presented in Figure \ref{fig:input_proj_1}: the transformer cannot reach the ID performance of OLS, and also performs significantly worse on OOD prompts.

To further characterize this behavior, Figure~\ref{fig:weight_proj_2} presents the error as the test weight $\hat{w}$ is formed by a convex combination:
\[
\hat{w} = t\,(P_A w) + (1-t)\,(P_Bw), \quad t \in [0,1].
\]

Again, we see nearly identical behaviour to that presented in Figure \ref{fig:input_proj_2}: the MSE becomes gradually larger as the vector's orthogonal subspace component becomes larger. Both of these results demonstrate the sensitivity of transformers to their pretraining distribution. These results also demonstrate that even distribution shifts in the weight vectors, which are never explicitly presented to the model, can also lead to poor generalization.

Moreover, we note the presence of the \textit{spectral signature} shown in Section~\ref{sec:spec_sig} under these training conditions as well. Notably, the singular-value spectrum for in distribution prompts has a sharp descent after the first two singular values, and more singular vectors are fixed across prompts. This supports the claim that this spectral signature occurs across many training distributions, and is not unique to a single trained task.

\begin{figure}[t]
    \centering
    \includegraphics[width=0.6\linewidth]{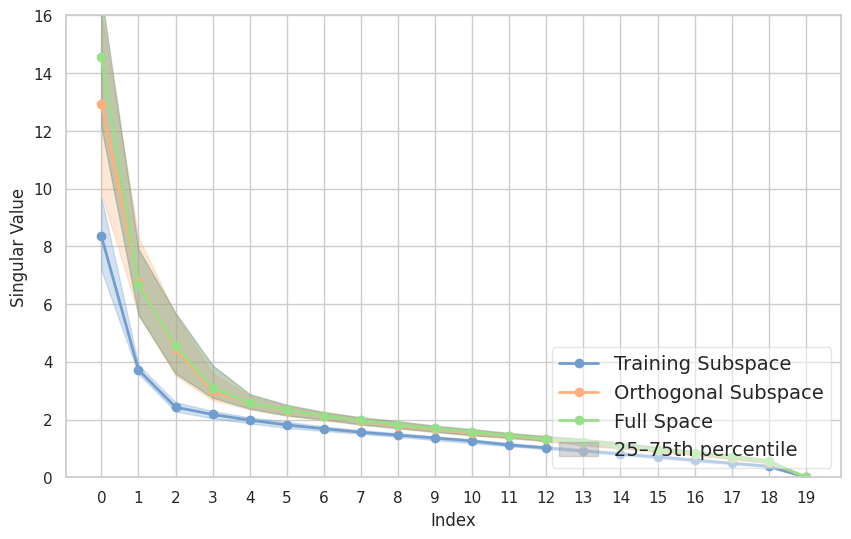}
    \caption{Singular-value spectrum of representations for the weight-space restricted transformer.}
    \label{fig:eigenspectrum_input_3}
\end{figure}

\begin{figure}[t]
    \centering
    \includegraphics[width=0.6\linewidth]{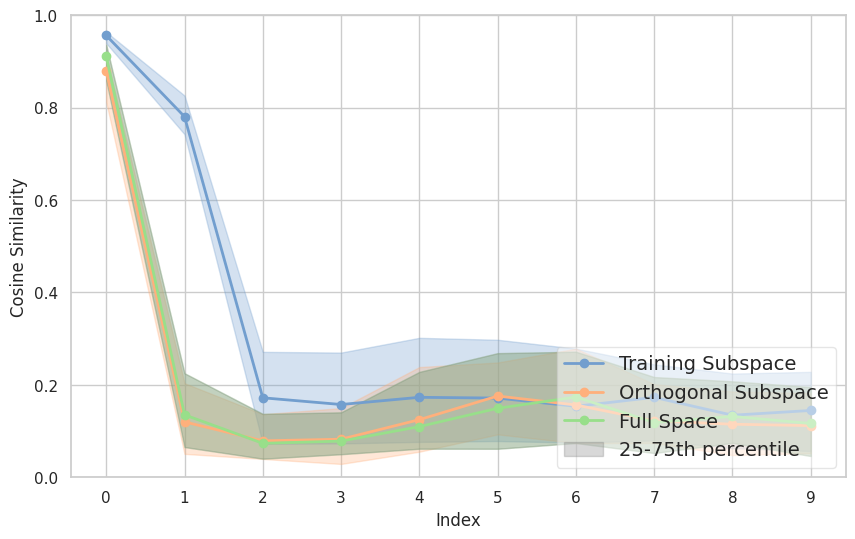}
    \caption{Cosine similarity between singular vectors of residual representations of the weight-space restricted transformer and its canonical set of singular vectors.}
    \label{fig:eigenspectrum_baseline_3}
\end{figure}

\begin{figure}[t]
    \centering
    \includegraphics[width=0.6\textwidth]{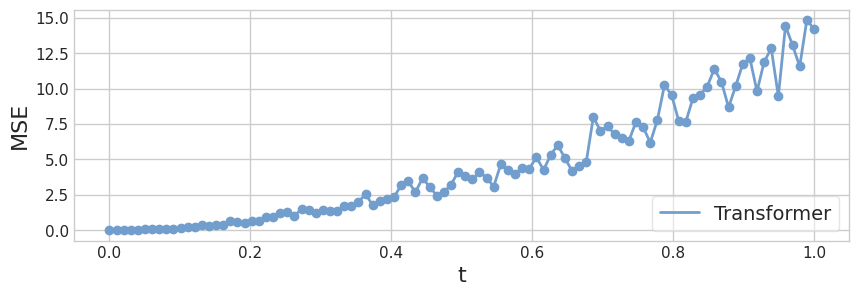}
    \hfill
    
    \caption{Prediction error as we blend an in‑distribution task with an OOD task. The horizontal axis is the convex combination coefficient $t$, and the vertical axis shows the resulting regression error.}
    \label{fig:weight_proj_2}
\end{figure}

\section{Gaussian Distribution Test for Out-of-Distribution Spectral Signatures}\label{sec:app_ood_detect}
Here we detail our statistical test that leverages the spectral signature to detect OOD data when a transformer is deployed. Let $V^{*\top}_{S_{||}}$ be the canonical singular vectors from the \textit{training subspace}. We start by calculating $C_p$ on a subset of the prompts from $S_\parallel$, $\widehat{S_\parallel}$. We collect the first two indices of this subset into the set $\widehat{C_{:2}} := \{C_{:2}| C_p := diag(V^{*\top}_{S_{||}}V_p), Z_p = U_p\Sigma_p V^\top_p, Z_p \in \widehat{S_\parallel}\}$   and fit a bivariate Gaussian to these values:

\begin{equation}
    \widehat{\mu_{:2}} := \frac{1}{|\widehat{C_{:2}}|} \sum_{C_{:2} \in |\widehat{C_{:2}}|}C_{:2}, \quad \widehat{\Sigma_{:2}}  := \frac{1}{|\widehat{C_{:2}}|} \sum_{C_{:2} \in |\widehat{C_{:2}}|} (C_{:2} -  \widehat{\mu_{:2}})(C_{:2} -  \widehat{\mu_{:2}})^\top.
\end{equation}

Once fit, we determine whether a new sample $\tilde{C_{:2}}$ is within the 95\% confidence region defined by the Gaussian with these moments.

The values presented in Table \ref{tab:ood_detector} are the percent of projections in a set $V_S$ that were within the specified confidence region, as determined using the following expression, where $\chi^2_{2}(0.95)$ is the quantile function for a Chi-squared distribution with two degrees of freedom:

\begin{equation}
    \frac{1}{|V_S|}\sum_{\tilde{C_{:2}} \in V_S}\mathbb{1}[(\tilde{C_{:2}} -  \widehat{\mu_{:2}})\widehat{\Sigma_{:2}}^{-1}(\tilde{C_{:2}} -  \widehat{\mu_{:2}})^\top \leq \chi^2_{2}(0.95)].
\end{equation}

We repeat this experiment 20 times and take the mean and standard deviation of this percent across trials.

\section{Preliminary Spectral Signatures in Language Models}\label{sec:app_LLM}

\begin{figure}[h]
    \centering
    \includegraphics[width=0.6\linewidth]{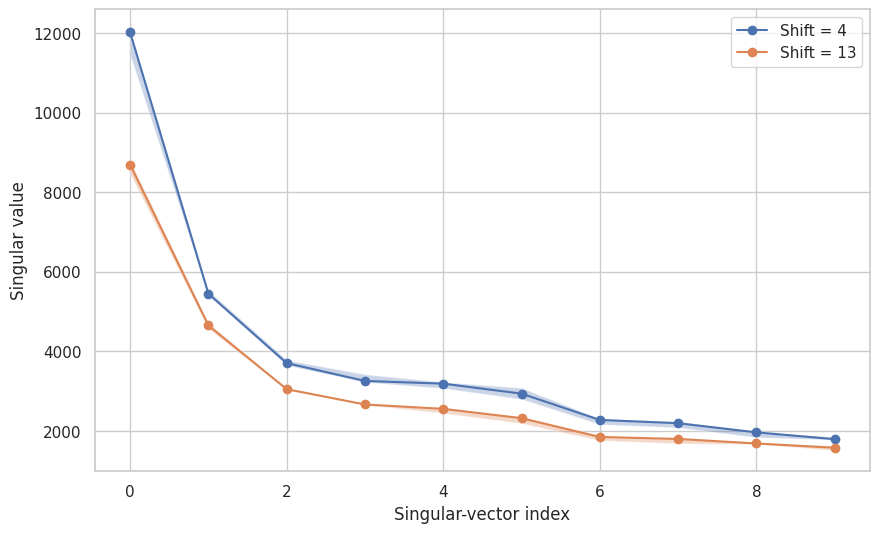}
    \caption{Comparison of singular-value spectrum between representations for $4$-shifted (OOD) prompts and $13$-shifted (in-distribution) prompts.}
    \label{fig:lang_spec}
\end{figure}

\begin{figure}[h]
    \centering
    \includegraphics[width=0.6\linewidth]{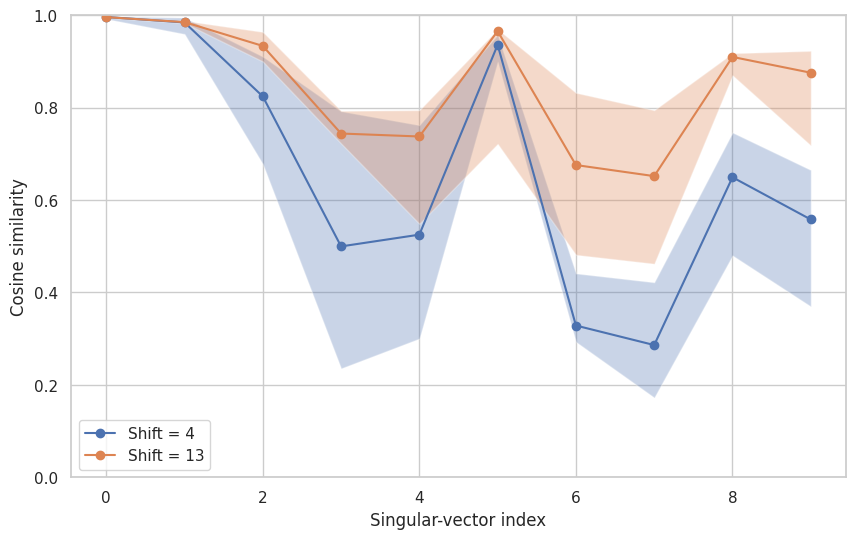}
    \caption{Comparison of cosine similarity between singular vectors of residual representations for $4$-shifted (OOD) prompts and $13$-shifted (in-distribution) prompts.}
    \label{fig:lang_cos}
\end{figure}

Building on the results from \citet{mccoy2023embers}, we extend our synthetic analysis to a real‐world task: decoding rotational Caesar ciphers.  

Our experimental setup uses the Qwen3-4B model \cite{qwen3}. Prompts are constructed according to the following template:

\begin{verbatim}
Here are some Caesar cipher examples (shift={shift}): {examples}.
Please decode the following cipher text: {test}
Output your final answer as: <answer>decoded_word</answer>
\end{verbatim}

Here, `\{shift\}' is the integer specifying the cipher rotation amount; `\{examples\}' is a comma‐separated list of example plaintext‐ciphertext pairs (e.g. uryyb → hello, ovt → big); `\{test\}' is the target text for decoding.

By contrasting a frequently seen shift (e.g.\ 13) with a rarer shift (e.g.\ 4), we probe how out‐of‐distribution inputs manifest in the model’s internal representations. With a batch of $150$ prompts per shift amount, the transformer achieves an accuracy of $43\%$ for $13$-shifted prompts, and $10\%$ for $4$-shifted prompts, demonstraing a generalization gap to OOD data. Figures \ref{fig:lang_spec} and \ref{fig:lang_cos} plot the singular value spectra and the alignment consistency of the top singular vectors for batches of 13‐shifted versus 4‐shifted prompts. Consistent with our synthetic results, the OOD (4‐shifted) prompts exhibit both higher initial singular values and reduced cosine similarity among singular vectors, whereas the in‐distribution (13‐shifted) prompts show lower‐magnitude spectra and stronger vector alignment.

\end{document}